\documentclass{article}

\usepackage{iclr2027_conference,times}
\iclrfinalcopy 

\usepackage[utf8]{inputenc}
\usepackage[T1]{fontenc}
\usepackage{hyperref}
\hypersetup{
  hidelinks,
  pdftitle={NGM: A Plug-and-Play Training-Free Memory Module for LLMs},
  pdfauthor={Yuwen Qu, Wenhui Dong, Camil Hamami, Ming Zhao, Chenyang Si, Caifeng Shan}
}
\usepackage{url}
\usepackage{booktabs}
\usepackage{placeins}
\usepackage{amsfonts}
\usepackage{amsmath}
\usepackage{nicefrac}
\usepackage{microtype}
\usepackage{xcolor}
\usepackage{array,colortbl}
\usepackage{tikz}
\usepackage{textcomp}

\usepackage{pifont}
\usepackage[most]{tcolorbox}
\usepackage{graphicx}
\usepackage{wrapfig}

\AddToHook{env/table/begin}{\setlength{\tabcolsep}{5.7pt}}

\usepackage{algorithm}
\usepackage{algpseudocode}

\usepackage{listings}

\lstdefinestyle{pypseudo}{
  language=Python,
  basicstyle=\ttfamily\footnotesize,
  columns=fullflexible,
  keepspaces=true,
  showstringspaces=false,
  breaklines=true,
  breakatwhitespace=true,
  frame=single,
  rulecolor=\color{black},
  captionpos=b,
  xleftmargin=0.5em,
  xrightmargin=0.5em,
  aboveskip=0.5em,
  belowskip=0.5em
}

\newcommand{\method}{NGM} 
\newcommand{\ngram}{N\text{-}gram}
\newcommand{\scaleparam}{\lambda}

\title{NGM: A Plug-and-Play Training-Free Memory Module for LLMs}

\author{\normalfont\makebox[\dimexpr\textwidth-2\tabcolsep\relax][c]{\begin{tabular}[t]{@{}c@{\hspace{2em}}c@{\hspace{2em}}c@{}}\textbf{Yuwen Qu}\textsuperscript{1,*} & \textbf{Wenhui Dong}\textsuperscript{1,*} & \textbf{Camil Hamami}\textsuperscript{2}\\[0.6em]\textbf{Ming Zhao}\textsuperscript{1} & \textbf{Chenyang Si}\textsuperscript{1} & \textbf{Caifeng Shan}\textsuperscript{1,\ensuremath{\dagger}}\\[0.8em]\multicolumn{3}{c}{\small \textsuperscript{1}Nanjing University \quad \textsuperscript{2}Tsinghua University}\\[0.5em]\multicolumn{3}{c}{\ttfamily\small cfshan@nju.edu.cn}\end{tabular}}}

\begin{document}

\maketitle
\fancyhead{}
\renewcommand{\headrulewidth}{0pt}

\ificlrfinal
\begingroup
\renewcommand{\thefootnote}{\fnsymbol{footnote}}
\footnotetext[1]{Equal contribution.}
\footnotetext[2]{Corresponding author.}
\endgroup\fi

\begin{abstract}

Recent studies have introduced memory-augmented architectures that decouple knowledge storage from neural computation, enabling more direct access to stored representations. However, existing approaches often rely on learned memory embeddings or explicit memory and retrieval components, requiring additional training or increasing architectural complexity. To address these limitations, we propose \textbf{N-gram Memory (NGM)}, a training-free, plug-and-play module that repurposes a backbone model's pretrained token embedding space as memory. NGM consists of a \textbf{Causal N-Gram Encoder} and a \textbf{Cosine-Gated Memory Injector}. The former constructs N-gram representations by directly aggregating pretrained token embeddings, while the latter uses cosine similarity with ReLU gating to adaptively control the amount of retrieved memory injected into contextual representations. Unlike conventional memory-augmented architectures, NGM requires neither additional embedding training nor a separately learned memory bank or external retrieval infrastructure.
We evaluate NGM on ten benchmarks across six language-model backbones spanning two model families and multiple scales, including four Qwen3 models (0.6B, 1.7B, 4B, and 8B) and two Llama-3.2-Instruct models (1B and 3B). NGM consistently improves the ten-task average across all six backbones. Notably, it improves \textbf{HumanEval by 4.26}, \textbf{HumanEval+ by 2.41}, and \textbf{GPQA-Diamond by 6.56} percentage points. Furthermore, NGM extends beyond text-only language models to vision-language models, with Qwen3-VL-2B achieving improvements across all five VLM benchmarks.\\\mbox{Project page: \url{https://pioneerqyw.github.io/ngm-project/}.}\end{abstract}

\section{Introduction}

Transformer-based large language models (LLMs)~\citep{vaswani2017attention,zheng2023less,guo2025fila,jiang2026ivy,zhu2024fila,zhao2026spinebench}
excel at contextual modeling and compositional reasoning, but not all
patterns encountered during language modeling require the same form of
computation. Named entities, repeated identifiers, units, terminology,
and formulaic phrases often exhibit stable local dependencies that could
potentially be recovered through inexpensive lookup rather than repeatedly
reconstructed by deep neural computation~\citep{erman2000idiom,constant2017multiword,brants2007large,liu2024infini,nguyen2024understanding}.
Standard Transformers, however, lack a native lookup primitive for such
local lexical and symbolic patterns, and therefore rely on attention and
feed-forward computation to reconstruct them at inference time
~\citep{wu2022memorizing,cheng2026conditional}.

Lookup-style memory provides a natural mechanism for separating static
pattern reuse from dynamic Transformer computation
~\citep{khandelwal2019generalization,wu2022memorizing}. Recent approaches
have explored this direction by introducing explicitly learned memory
representations. Engram~\citep{cheng2026conditional}, for example,
formulates conditional memory using learned $N$-gram lookup tables and
context-dependent gating, while embedding-scaling methods increase memory
capacity through additional token-level or $N$-gram embedding parameters
~\citep{yu2026scaling,tseng20263,Liu2026ScalingEO,ding2026meki}.
These results show that local lookup can aid language models, but typically relies on separately learned memory representations, adding parameters, training, and sometimes dedicated storage or retrieval infrastructure.

This observation raises a more fundamental question:
\begin{quote}
\centering
\emph{Can a pretrained LLM embedding space serve as local memory without retraining or a separate memory table?}
\end{quote}

We challenge the assumption that useful memory representations must be
separately learned. Our central insight is to reinterpret the pretrained
token embedding space not merely as an input representation, but as a
\emph{non-parametric memory substrate}. If nearby tokens form stable
lexical, symbolic, or phrase-level patterns, their pretrained embeddings
may be aggregated into local $N$-gram representations that provide
reusable signals without introducing new memory parameters. Specifically,
we construct causal $N$-gram representations by averaging pretrained token
embeddings within local trailing windows. This simple operation converts
the backbone's existing embedding space into a lightweight local-memory
representation without learning a separate $N$-gram embedding space.

\begin{wrapfigure}{r}{0.5\textwidth}
    \centering
\includegraphics[width=\linewidth]{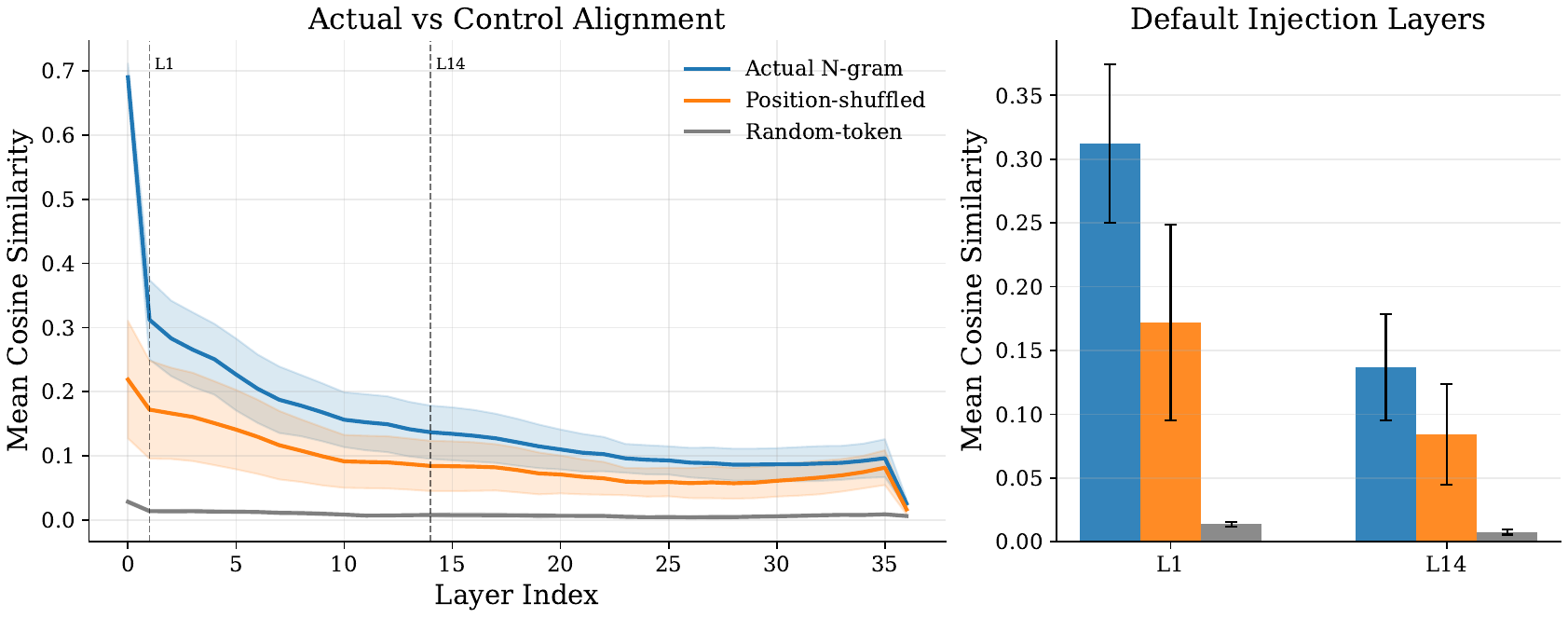}
    \caption{Alignment between hidden states and aggregated $N$-gram embeddings
    in the Qwen3-8B model.}\label{fig:mech_alignment_controls}
\end{wrapfigure}

A key question, however, is whether these embedding-derived $N$-gram
representations are compatible with the contextual representation space
of the decoder. Figure~\ref{fig:mech_alignment_controls} provides evidence
that they are. Across model depth in Qwen3-8B, aggregated $N$-gram
embeddings exhibit substantially stronger cosine alignment with hidden
states than both position-shuffled $N$-gram controls and random-token
controls. At the two default injection layers, the mean cosine
similarities are 0.312 and 0.137 for the actual $N$-gram representations,
compared with 0.172 and 0.084 for shuffled controls and 0.014 and 0.008
for random controls, respectively. These results suggest that the
pretrained embedding space preserves local $N$-gram structure that is
measurably compatible with contextual representations, motivating a
similarity-based mechanism for selectively injecting such memory signals.

\begin{figure*}[t]
\centering
\includegraphics[width=1\textwidth]{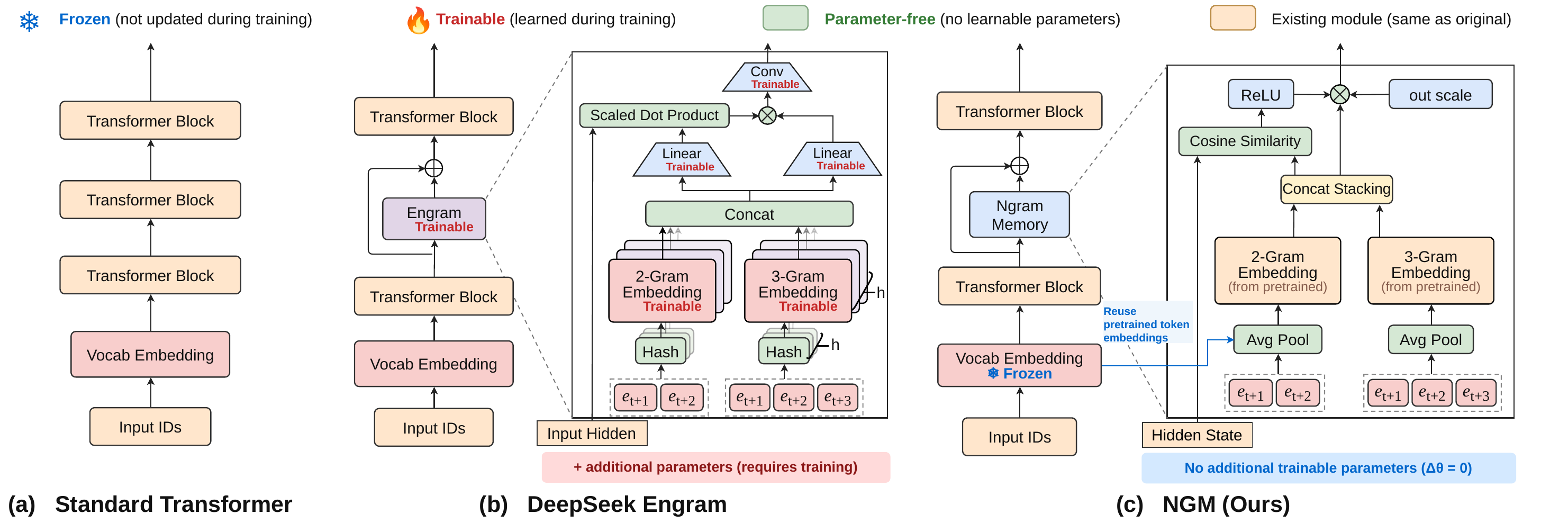}
    \caption{Overview of \method{}. 
\textbf{(a)} Standard Transformer without an explicit N-gram memory.
\textbf{(b)} DeepSeek Engram introduces trainable N-gram embeddings and additional learned modules.
\textbf{(c)} NGM reuses pretrained vocabulary embeddings to construct N-gram representations through parameter-free pooling and similarity-based retrieval, introducing no additional trainable parameters ($\Delta\theta = 0$).}
    \label{fig:framework}
\end{figure*}

Motivated by this observation, we propose
\textbf{\method{} (N-gram Memory)}, a training-free, plug-and-play memory
module for frozen decoder-only LLMs, as illustrated in Figure~\ref{fig:framework}. The central design principle of
\method{} is to reuse the backbone's pretrained representation space
rather than introducing a separately learned memory space. \method{} consists of two non-parametric
components: a \textbf{Causal \ngram{} Encoder} and a
\textbf{Cosine-Gated Memory Injector}. Given an input sequence, the Causal
\ngram{} Encoder constructs causal multi-scale $N$-gram representations by
aggregating pretrained token embeddings over local trailing windows,
capturing local patterns at different granularities without learning
additional memory entries. The Cosine-Gated Memory Injector then measures
the cosine similarity between these $N$-gram representations and decoder
hidden states, applies a ReLU-filtered gate, and injects the resulting
memory signal through a scaled residual connection. Consequently, only
positively aligned local-memory signals contribute to the contextual
representation, while unrelated or negatively aligned signals are
suppressed.
This design has two complementary implications. 
Practically, \method{} can be added to a pretrained LLM without updating the backbone, learning a separate embedding space, or using a memory bank or external retrieval system. Conceptually, it offers a simple framework for testing whether pretrained representations contain reusable local-memory structure.

We evaluate \method{} on ten benchmarks across six language-model
backbones spanning two model families and multiple scales: four Qwen3
models (0.6B, 1.7B, 4B, and 8B) and two Llama-3.2-Instruct models (1B and
3B). Across all six backbones, NGM variants consistently improve the
unweighted ten-task average over their corresponding base models. The improvements are
particularly pronounced on code generation and complex reasoning,
including gains of 4.26 points on HumanEval, 2.41 points on HumanEval+,
and 6.56 points on GPQA-Diamond. We further test the same design beyond
text-only language models by applying \method{} only to the language
decoder of Qwen3-VL-2B, leaving the visual encoder unchanged. NGM improves
performance across all five evaluated VLM benchmarks.

In summary, our contributions are threefold:
(1) we show that a pretrained token embedding space can serve as a
non-parametric substrate for local $N$-gram memory, challenging the
assumption that memory representations must be separately learned;
(2) we introduce \method{}, a training-free and plug-and-play memory
module that combines causal multi-scale $N$-gram aggregation with
cosine-gated memory injection; and
(3) we provide extensive empirical evaluation across six LLM backbones,
ten benchmarks, and a vision-language model, demonstrating consistent
improvements across model families and scales and substantial gains in
code generation and complex reasoning.

\section{Related Work}
\label{sec:related}

\textbf{Conditional memory and learned lookup representations.}
Recent work augments language models with learned lookup representations to
increase capacity or enable conditional access to stored patterns. SCONE
\citep{yu2026scaling} introduces an auxiliary Transformer for producing
contextualized $f$-gram embeddings, while L$^3$
\citep{tseng20263}, LongCat-Flash-Lite
\citep{Liu2026ScalingEO}, and MeKi
\citep{ding2026meki} explore token- or $N$-gram-level memory through learned
embeddings, conditional routing, or memory experts. Engram
\citep{cheng2026conditional} is most closely related to our work. It combines
hashed $N$-gram lookup, context-aware gating, and sparsity allocation to
provide conditional memory at intermediate layers. Qwen3.8-Flash-Next
\citep{qiu2026design} similarly uses hashed $N$-gram embeddings with
contextual gating.
These methods demonstrate the utility of lookup-based memory, but they
introduce a separately learned memory space and require additional training.
Several also rely on specialized routing, caching, host-memory prefetching, or
offloading mechanisms. As a result, their performance improvements are coupled
with increased model capacity and system complexity. In contrast, \method{}
reuses the backbone's existing token embedding space and constructs causal
multi-scale $N$-gram representations at inference time, without additional
training or learned memory parameters.

\textbf{Retrieval-based memory.}
A complementary line of work retrieves stored model representations, as in
kNN-LMs and Memorizing Transformers
\citep{khandelwal2019generalization,wu2022memorizing}, or text from external
corpora, as in REALM and RETRO
\citep{guu2020retrieval,borgeaud2022improving}. These methods provide access to
information beyond the current computation but require explicit datastores,
indexing, and retrieval procedures. \method{} instead operates entirely within
the frozen backbone and uses its pretrained embeddings as an internal,
non-parametric source of local memory.

\section{Methodology}
\label{sec:method}

\method{} is a training-free, plug-and-play memory module for frozen
decoder-only language models. It is based on the following hypothesis:

\emph{if the pretrained token embedding space preserves reusable local
regularities, then these regularities can be exposed as memory without learning
a separate memory representation.}

This hypothesis leads to a two-stage design. First, the
\textbf{\emph{Causal \ngram{} Encoder}} constructs local-memory candidates by
aggregating pretrained token embeddings over causal windows of multiple
scales. Second, the \textbf{\emph{Cosine-Gated Memory Injector}} uses the current
decoder state to measure the compatibility of each candidate and writes only
positively aligned signals into the residual stream.

This design separates two roles that are often entangled in
memory-augmented architectures: the encoder specifies \emph{what local
information is available}, while the injector determines \emph{when that
information should affect computation}. All backbone parameters remain frozen,
and \method{} introduces no separately learned memory table, external
datastore, or trainable projection.

\subsection{Problem Setup and Zero-Parameter Memory Reuse}
\label{sec:setup}

Let
$
\boldsymbol{X}=\{x_1,\ldots,x_T\}
$
denote an input token sequence of length $T$. Let
$\boldsymbol{E}\in\mathbb{R}^{V\times d}$ be the frozen token embedding matrix
of the backbone, where $V$ is the vocabulary size and $d$ is the embedding
dimension. The embedding of token $x_t$ is
$
\boldsymbol{e}_t=\boldsymbol{E}[x_t]\in\mathbb{R}^{d}.
$

Unlike conventional memory-augmented models, \method{} does not introduce a
new parameterized memory space. Instead, it treats local transformations of
$\boldsymbol{E}$ as memory candidates and uses the decoder hidden state as a
context-dependent compatibility query. The resulting memory operation is
therefore zero-parameter with respect to the memory mechanism itself.

\subsection{Causal Multi-Scale Memory Construction}
\label{sec:causal_encoder}

Let $\mathcal{N}$ denote a predefined set of window sizes. For each
$n\in\mathcal{N}$, we construct a causal local representation using the current
token and its preceding $n-1$ tokens. To preserve the sequence length, we
prepend $n-1$ zero vectors:
\begin{equation}
\tilde{\boldsymbol{e}}_t =
\begin{cases}
\boldsymbol{0}, & 2-n\leq t\leq 0,\\
\boldsymbol{e}_t, & 1\leq t\leq T.
\end{cases}
\end{equation}
The causal $N$-gram representation at position $t$ is
\begin{equation}
\boldsymbol{g}_{t,n}
=
\frac{1}{n}
\sum_{k=0}^{n-1}
\tilde{\boldsymbol{e}}_{t-k}.
\label{eq:causal_ngram}
\end{equation}

The averaging operation is intentionally parameter-free. It preserves the
backbone dimension, introduces no additional memory entries, and exposes local
patterns at multiple context scales. More importantly, it defines a common
representation space in which memory construction and contextual compatibility
can be evaluated without learned alignment layers.

For each position, the multi-scale memory candidates are stacked as
\begin{equation}
\boldsymbol{G}_t
=
\left[
\boldsymbol{g}_{t,n}
\right]_{n\in\mathcal{N}}
\in\mathbb{R}^{|\mathcal{N}|\times d}.
\label{eq:memory_candidates}
\end{equation}

The construction is causal because every
$\boldsymbol{g}_{t,n}$ depends only on tokens up to position $t$. Thus,
$\boldsymbol{G}_t$ can be computed during both full-sequence prefill and
incremental autoregressive decoding.

\subsection{Cosine-Gated Memory Injection}
\label{sec:injection}

The Causal \ngram{} Encoder provides candidate local memories, but not every
candidate is useful for every context. The Cosine-Gated Memory Injector
addresses this selection problem without introducing a learned controller.

Let $\boldsymbol{h}_t^l\in\mathbb{R}^{d}$ be the hidden state at position $t$
and layer $l$. We measure the compatibility between the hidden state and each
candidate using cosine similarity:
\begin{equation}
s_{t,n}
=
\cos\left(
\boldsymbol{h}_t^l,\boldsymbol{g}_{t,n}
\right)
=
\frac{
\left\langle
\boldsymbol{h}_t^l,\boldsymbol{g}_{t,n}
\right\rangle
}{
\left\|\boldsymbol{h}_t^l\right\|_2
\left\|\boldsymbol{g}_{t,n}\right\|_2
}.
\label{eq:cosine_compatibility}
\end{equation}

The score serves as a geometric compatibility test between the contextual
representation and the local-memory candidate. We then apply a ReLU gate:
\begin{equation}
\alpha_{t,n}
=
\max(0,s_{t,n}).
\label{eq:positive_gate}
\end{equation}

The ReLU is important for the intended memory behavior. Positively aligned
candidates can reinforce the current representation, whereas negatively
aligned candidates are treated as incompatible and are prevented from
contributing to the update.
The memory signal injected at layer $l$ is
\begin{equation}
\boldsymbol{m}_t^l
=
\sum_{n\in\mathcal{N}}
\alpha_{t,n}
\boldsymbol{g}_{t,n}.
\label{eq:gated_memory}
\end{equation}
We write this signal into the residual stream using
\begin{equation}
\boldsymbol{h}_t^{l'}
=
\boldsymbol{h}_t^l
+
\scaleparam_l\boldsymbol{m}_t^l,
\label{eq:memory_write}
\end{equation}
where $\scaleparam_l$ is a fixed injection scale for layer $l$.
Equations~\eqref{eq:cosine_compatibility}--\eqref{eq:memory_write} define a
content-addressed residual write. The pretrained embedding space provides the
candidate signal, cosine similarity determines its contextual compatibility,
and the residual connection controls how the selected signal enters the
decoder computation. No learned key-value projection, retrieval index, or additional memory
parameters are required. The complete algorithmic implementation is provided in Algorithm~\ref{alg:ngm_simple}.

\begin{algorithm}[ht]
\caption{\method{}: N-gram Memory (\emph{Causal \ngram{} Encoder} + \emph{Cosine-Gated Memory Injector})}
\label{alg:ngm_simple}
\footnotesize
\begin{algorithmic}[1]
\Require Token IDs $\boldsymbol{X}=\{x_1,\dots,x_T\}$; hidden states $\boldsymbol{H}^l=\{\boldsymbol{h}_t^l\}_{t=1}^{L}$; backbone token embedding matrix $\boldsymbol{E}\in\mathbb{R}^{V\times d}$; \ngram{} sizes $\mathcal{N}$; output scale $\scaleparam$; boolean use\_relu.
\Ensure Updated hidden states $\boldsymbol{H}^{l'}=\{\boldsymbol{h}_t^{l'}\}_{t=1}^{L}$.
\Statex \textit{\% --- Causal \ngram{} Encoder (Eqs.~1--4) ---}
\State $\boldsymbol{G} \leftarrow \textsc{CausalNgramEncoder}(\boldsymbol{X}, \boldsymbol{E}, \mathcal{N})_{T-L+1:T}$ \Comment{$\boldsymbol{G}\!\in\!\mathbb{R}^{L\times|\mathcal{N}|\times d}$, aligned to last $L$ positions}
\Statex \textit{\% --- Cosine-Gated Memory Injector ---}
\State $\hat{\boldsymbol{H}}^l \leftarrow \text{L2Norm}(\boldsymbol{H}^l)$ \Comment{Row-wise $\ell_2$ normalization;\; $\hat{\boldsymbol{H}}^l\!\in\!\mathbb{R}^{L\times d}$}
\State $\hat{\boldsymbol{G}} \leftarrow \text{L2Norm}(\boldsymbol{G})$ \Comment{Normalize along $d$;\; $\hat{\boldsymbol{G}}\!\in\!\mathbb{R}^{L\times|\mathcal{N}|\times d}$}
\State $S_{t,n} \leftarrow \sum_{j} \hat{H}^l_{t,j}\, \hat{G}_{t,n,j}$ \Comment{Position-wise cosine similarity;\; $\boldsymbol{S}\!\in\!\mathbb{R}^{L\times|\mathcal{N}|}$}
\If{use\_relu}
    \State $\boldsymbol{S} \leftarrow \max(\boldsymbol{0},\, \boldsymbol{S})$ \Comment{Suppress negatively aligned entries}
\EndIf
\State $M_{t,j} \leftarrow \sum_{n} S_{t,n}\, G_{t,n,j}$ \Comment{Gated weighted sum over \ngram{} scales;\; $\boldsymbol{M}\!\in\!\mathbb{R}^{L\times d}$}
\State $\boldsymbol{H}^{l'} \leftarrow \boldsymbol{H}^l + \scaleparam\, \boldsymbol{M}$ \Comment{Scaled residual injection}
\State \Return $\boldsymbol{H}^{l'}$
\end{algorithmic}
\end{algorithm}

\subsection{Layer Placement and Cached Decoding}
\label{sec:layer_cache}

We apply the injector after the MLP block in a small set of selected decoder
layers. Let
$\mathcal{L}_{\mathrm{inj}}$ denote the injection-layer set. The original
self-attention and MLP parameters are unchanged; \method{} only modifies the
layer representation through Eq.~\eqref{eq:memory_write}. The layer IDs and
injection scales are fixed hyperparameters, whose default values are reported
in Appendix Table~\ref{tab:appendix_ngm_settings}.
Motivated by the layer-sensitivity analysis of Engram~\citep{cheng2026conditional},
we use a sparse placement as a shared practical default.

During prefill, the encoder constructs $\boldsymbol{G}_t$ for all input
positions. During autoregressive decoding with a KV cache, the representation
at position $t$ depends only on the current token and at most the preceding
$\max(\mathcal{N})-1$ token embeddings. We therefore maintain a short causal
embedding cache and update the memory candidates incrementally.

If the active hidden-state segment contains $L$ positions, we align the
corresponding memory candidates by selecting the final $L$ positions:
\begin{equation}
\boldsymbol{G}_{T-L+1:T}
=
\left[
\boldsymbol{G}_{T-L+1},\ldots,\boldsymbol{G}_{T}
\right].
\label{eq:cache_alignment}
\end{equation}

This alignment preserves the causal correspondence between hidden states and
memory candidates while avoiding recomputation over the entire prefix.

\begingroup
\begin{table*}[t]
\centering
\caption{Performance across six backbones and ten benchmarks
(\%; $\uparrow$).
For each backbone and benchmark, the baseline and NGM variants were evaluated
under the same decoding configuration and run protocol; the table reports the
mean and sample standard deviation across three runs for NGM.
$\Delta$ denotes the difference between the NGM mean
and the baseline score, in percentage points.
Avg is the unweighted average across the ten benchmarks,
computed before rounding.}
\label{tab:main_results}
\label{tab:aggregate_mean_std}
\vspace{5pt}
\scriptsize
\setlength{\tabcolsep}{0pt}
\renewcommand{\arraystretch}{1.18}
\newlength{\ngmDataWidth}
\setlength{\ngmDataWidth}{\dimexpr(\linewidth-107pt)/10\relax}
\newcommand{\ngmDelta}[2]{\cellcolor[HTML]{#1}\rule[-1.6pt]{0pt}{10.8pt}#2}
\newcommand{\ngmStat}[2]{\mbox{#1{\fontsize{4.8}{5.5}\selectfont\textpm#2}}}
\arrayrulecolor{black!55}
\begin{tabular}{@{}p{45pt}@{\hspace{2pt}}p{31pt}@{\hspace{1pt}}*{4}{>{\centering\arraybackslash}p{\ngmDataWidth}@{}}>{\centering\arraybackslash}p{\ngmDataWidth}@{\hspace{1pt}\vrule width .3pt\hspace{1pt}}*{4}{>{\centering\arraybackslash}p{\ngmDataWidth}@{}}>{\centering\arraybackslash}p{\ngmDataWidth}@{\hspace{1pt}\vrule width .3pt\hspace{1pt}}>{\centering\arraybackslash}p{23pt}@{}}
\toprule
& & \multicolumn{5}{c}{\textbf{Mathematics and code generation}} & \multicolumn{5}{c}{\textbf{Knowledge, truthfulness, and reasoning}} & \\
\cmidrule(lr){3-7}\cmidrule(lr){8-12}
\textbf{Model} & & \textbf{GSM8K} & \textbf{MATH} & \textbf{HE} & \textbf{HE+} & \textbf{LCB} & \mbox{{\fontsize{6.3}{7}\selectfont\textbf{MMLU-R}}} & \textbf{GPQA} & \textbf{TQA} & \textbf{PopQA} & \textbf{BBH} & \textbf{Avg} \\
\midrule
\textbf{Qwen3-0.6B} & Baseline & 62.09 & 50.20 & 33.54 & 29.27 & 4.19 & 45.63 & 25.25 & 12.61 & 10.53 & 31.30 & 30.46 \\
 & +NGM & \ngmStat{\textbf{62.75}}{0.51} & \ngmStat{\textbf{52.40}}{1.56} & \ngmStat{\textbf{36.18}}{2.53} & \ngmStat{\textbf{31.51}}{2.31} & \ngmStat{\textbf{6.59}}{0.00} & \ngmStat{\textbf{46.04}}{0.29} & \ngmStat{\textbf{26.94}}{2.49} & \ngmStat{\textbf{13.91}}{0.14} & \ngmStat{\textbf{10.76}}{0.03} & \ngmStat{\textbf{31.70}}{0.13} & \textbf{31.88} \\
 & $\Delta$ & \ngmDelta{EAF5F0}{+0.66} & \ngmDelta{B9DECE}{+2.20} & \ngmDelta{ABD8C4}{+2.64} & \ngmDelta{B8DECD}{+2.24} & \ngmDelta{B3DBCA}{+2.40} & \ngmDelta{F2F9F6}{+0.41} & \ngmDelta{C9E6DA}{+1.69} & \ngmDelta{D6ECE2}{+1.30} & \ngmDelta{F8FCFA}{+0.23} & \ngmDelta{F2F9F6}{+0.40} & \ngmDelta{D2EAE0}{+1.42} \\
\midrule
\textbf{Qwen3-1.7B} & Baseline & 80.97 & \textbf{73.80} & 59.15 & 57.32 & 8.98 & 66.60 & 26.77 & 52.02 & 13.70 & 54.71 & 49.40 \\
 & +NGM & \ngmStat{\textbf{83.19}}{0.94} & \ngmStat{73.60}{0.53} & \ngmStat{\textbf{63.41}}{1.22} & \ngmStat{\textbf{59.73}}{0.61} & \ngmStat{\textbf{9.78}}{0.35} & \ngmStat{\textbf{67.09}}{0.37} & \ngmStat{\textbf{33.33}}{1.82} & \ngmStat{\textbf{52.51}}{0.21} & \ngmStat{13.70}{0.06} & \ngmStat{\textbf{54.87}}{0.36} & \textbf{51.12} \\
 & $\Delta$ & \ngmDelta{B9DECE}{+2.22} & \ngmDelta{FEFAFA}{-0.20} & \ngmDelta{78C0A1}{+4.26} & \ngmDelta{B3DBCA}{+2.41} & \ngmDelta{E6F3ED}{+0.80} & \ngmDelta{EFF8F4}{+0.49} & \ngmDelta{41A67A}{+6.56} & \ngmDelta{EFF8F4}{+0.49} & \ngmDelta{F7F7F7}{+0.00} & \ngmDelta{FAFDFB}{+0.16} & \ngmDelta{C9E5D9}{+1.72} \\
\midrule
\textbf{Qwen3-4B} & Baseline & 91.58 & 83.20 & 78.66 & 73.78 & 18.56 & 79.21 & 42.42 & 51.29 & 15.80 & 79.47 & 61.40 \\
 & +NGM & \ngmStat{\textbf{91.81}}{0.20} & \ngmStat{\textbf{85.07}}{0.58} & \ngmStat{\textbf{81.51}}{0.35} & \ngmStat{\textbf{75.81}}{0.70} & \ngmStat{\textbf{19.56}}{0.69} & \ngmStat{\textbf{79.83}}{0.38} & \ngmStat{\textbf{47.47}}{1.34} & \ngmStat{\textbf{52.10}}{0.49} & \ngmStat{\textbf{15.88}}{0.02} & \ngmStat{\textbf{80.03}}{0.35} & \textbf{62.91} \\
 & $\Delta$ & \ngmDelta{F8FCFA}{+0.23} & \ngmDelta{C4E3D6}{+1.87} & \ngmDelta{A5D5C0}{+2.85} & \ngmDelta{BFE1D2}{+2.03} & \ngmDelta{DFF0E9}{+1.00} & \ngmDelta{EBF6F1}{+0.62} & \ngmDelta{5FB48F}{+5.05} & \ngmDelta{E5F3ED}{+0.81} & \ngmDelta{FCFEFD}{+0.08} & \ngmDelta{EDF7F3}{+0.56} & \ngmDelta{CFE9DE}{+1.51} \\
\midrule
\textbf{Qwen3-8B} & Baseline & 93.18 & 84.00 & 83.54 & 82.32 & 19.76 & 80.72 & 42.93 & 72.95 & 19.98 & 82.94 & 66.23 \\
 & +NGM & \ngmStat{\textbf{93.66}}{0.29} & \ngmStat{\textbf{85.00}}{0.92} & \ngmStat{\textbf{85.17}}{0.93} & \ngmStat{82.32}{0.61} & \ngmStat{\textbf{21.16}}{0.69} & \ngmStat{\textbf{80.98}}{0.20} & \ngmStat{\textbf{47.14}}{2.87} & \ngmStat{\textbf{73.19}}{0.13} & \ngmStat{\textbf{20.28}}{0.05} & \ngmStat{\textbf{83.25}}{0.17} & \textbf{67.21} \\
 & $\Delta$ & \ngmDelta{F0F8F4}{+0.48} & \ngmDelta{DFF0E9}{+1.00} & \ngmDelta{CBE7DB}{+1.63} & \ngmDelta{F7F7F7}{+0.00} & \ngmDelta{D3EAE0}{+1.40} & \ngmDelta{F7FBF9}{+0.26} & \ngmDelta{7AC1A2}{+4.21} & \ngmDelta{F7FBFA}{+0.24} & \ngmDelta{F6FBF8}{+0.30} & \ngmDelta{F5FAF8}{+0.31} & \ngmDelta{E0F0E9}{+0.98} \\
\midrule
\textbf{Llama-1B} & Baseline & 35.71 & 20.80 & 25.61 & 25.61 & 0.00 & 35.44 & 19.19 & 33.90 & 11.42 & 23.16 & 23.08 \\
 & +NGM & \ngmStat{\textbf{37.15}}{0.46} & \ngmStat{\textbf{22.73}}{0.46} & \ngmStat{\textbf{29.68}}{0.93} & \ngmStat{\textbf{27.03}}{1.41} & \ngmStat{\textbf{0.80}}{0.35} & \ngmStat{\textbf{36.74}}{0.53} & \ngmStat{\textbf{20.88}}{2.59} & \ngmStat{\textbf{35.25}}{0.43} & \ngmStat{\textbf{11.52}}{0.02} & \ngmStat{\textbf{23.61}}{0.20} & \textbf{24.54} \\
 & $\Delta$ & \ngmDelta{D1EADF}{+1.44} & \ngmDelta{C2E2D4}{+1.93} & \ngmDelta{7EC3A5}{+4.07} & \ngmDelta{D2EAE0}{+1.42} & \ngmDelta{E6F3ED}{+0.80} & \ngmDelta{D6ECE2}{+1.30} & \ngmDelta{C9E6DA}{+1.69} & \ngmDelta{D4EBE1}{+1.35} & \ngmDelta{FCFEFD}{+0.10} & \ngmDelta{F1F8F5}{+0.45} & \ngmDelta{D1E9DF}{+1.46} \\
\midrule
\textbf{Llama-3B} & Baseline & 69.98 & 36.60 & 48.78 & 46.34 & 3.59 & 62.19 & 26.26 & 58.75 & 17.28 & 42.11 & 41.19 \\
 & +NGM & \ngmStat{\textbf{70.38}}{0.84} & \ngmStat{\textbf{37.53}}{1.10} & \ngmStat{\textbf{50.61}}{1.22} & \ngmStat{\textbf{47.36}}{0.35} & \ngmStat{\textbf{6.59}}{0.60} & \ngmStat{\textbf{62.72}}{0.16} & \ngmStat{\textbf{29.63}}{1.77} & \ngmStat{\textbf{60.18}}{0.43} & \ngmStat{\textbf{17.40}}{0.03} & \ngmStat{\textbf{42.68}}{0.15} & \textbf{42.51} \\
 & $\Delta$ & \ngmDelta{F2F9F6}{+0.40} & \ngmDelta{E2F1EA}{+0.93} & \ngmDelta{C5E4D6}{+1.83} & \ngmDelta{DFF0E8}{+1.02} & \ngmDelta{A0D3BD}{+3.00} & \ngmDelta{EEF7F3}{+0.53} & \ngmDelta{94CDB4}{+3.37} & \ngmDelta{D2EADF}{+1.43} & \ngmDelta{FBFDFC}{+0.12} & \ngmDelta{EDF7F2}{+0.57} & \ngmDelta{D5EBE2}{+1.32} \\
\bottomrule
\end{tabular}
\definecolor{ngmLegendRed}{HTML}{D9484B}
\definecolor{ngmLegendGreen}{HTML}{2B6D49}
\vspace{5pt}
{\fontsize{6}{7}\selectfont
\noindent\hbox to \linewidth{\hfil
$\Delta$ (NGM $-$ Base, pp)\hspace{7pt}
\begin{tikzpicture}[x=1pt,y=1pt,baseline=-7pt]
  \shade[left color=ngmLegendRed,right color=white] (0,0) rectangle (60,6);
  \shade[left color=white,right color=ngmLegendGreen] (60,0) rectangle (120,6);
  \node[anchor=north,font=\fontsize{5}{6}\selectfont] at (0,-1.5) {$-6$};
  \node[anchor=north,font=\fontsize{5}{6}\selectfont] at (30,-1.5) {$-3$};
  \node[anchor=north,font=\fontsize{5}{6}\selectfont] at (60,-1.5) {$0$};
  \node[anchor=north,font=\fontsize{5}{6}\selectfont] at (90,-1.5) {$+3$};
  \node[anchor=north,font=\fontsize{5}{6}\selectfont] at (120,-1.5) {$+6$};
\end{tikzpicture}
\hspace{12pt}\vrule width .3pt height 15pt depth 1pt\hspace{7pt}
\parbox[c]{120pt}{\fontsize{6}{7}\selectfont Green: improvement\enspace|\enspace Red: degradation\\Color encodes the difference, not the absolute score.}
\hfil}}
\vspace{-12pt}
\end{table*}
\endgroup

\subsection{Computational and Parameter Complexity}
\label{sec:complexity}

For a sequence of length $T$, embedding dimension $d$, and window-size set
$\mathcal{N}$, the causal memory construction requires
$O\left(T|\mathcal{N}|d\right)$
additional computation during prefill.
Let
$K=|\mathcal{L}_{\mathrm{inj}}|$ be the number of injection layers. The
cosine scoring, gating, and multi-scale aggregation add
$O\left(KT|\mathcal{N}|d\right)$
computation during prefill. With incremental embedding caching, the additional
cost per autoregressive decoding step is
$O\left(K|\mathcal{N}|d\right)$,
which does not grow with the prefix length.
The method introduces no trainable parameters. During streaming decoding, the
additional state consists only of the recent token embeddings needed to update
the causal windows, giving an auxiliary memory cost of
$O\left(\max(\mathcal{N})d\right)$.

\section{Experiments}
\label{sec:experiments}

\subsection{Experimental setup}
 
\paragraph{Setup.}
We evaluate \method{} on six backbones: Qwen3 at 0.6B, 1.7B,
4B, and 8B~\citep{yang2025qwen3}, and Llama-3.2-Instruct
at 1B and 3B~\citep{grattafiori2024llama,meta2024llama32}.
For each backbone, we compare the original checkpoint with its
\method{}-augmented counterpart, keeping all backbone weights frozen.
For each backbone, the \method{} configuration is fixed across tasks;
the output scales and insertion layers are reported in
Appendix Table~\ref{tab:appendix_ngm_settings}.
We use EvalScope~\citep{evalscope_2024} for these evaluations.
For each benchmark, the baseline and \method{} share identical
inference settings, including any benchmark-specific overrides.
The default settings are a maximum context length of
$32{,}768$ tokens, temperature $=0.7$, top-$p=0.8$,
top-$k=20$, and presence penalty $=1.5$.

\paragraph{Benchmarks.}
We evaluate mathematical reasoning on GSM8K~\citep{cobbe2021training}
and MATH500~\citep{hendrycks2021measuring}; code generation on
HumanEval~\citep{chen2021evaluating},
HumanEval+~\citep{liu2023evalplus}, and LiveCodeBench
v5~\citep{jain2025livecodebench}; knowledge and reasoning on
MMLU-Redux~\citep{hendrycks2020measuring,gema2025we},
GPQA-Diamond~\citep{rein2023gpqa}, and
BBH~\citep{suzgun2023challenging}; and truthfulness and factual
recall on TruthfulQA~\citep{lin2022truthfulqa} and
PopQA~\citep{mallen2023not}.

\subsection{Main results}

Table~\ref{tab:main_results} reports the results across six backbones and ten
benchmarks. NGM improves the unweighted ten-task average for every model, with
gains of 0.98--1.72 percentage points across both the Qwen3 and
Llama-3.2-Instruct families. The improvement is not monotonic with model size:
Qwen3-1.7B obtains the largest average gain (+1.72), whereas Qwen3-8B gains
+0.98. This suggests that NGM acts as a complementary signal to the frozen
backbone rather than providing a fixed improvement that scales uniformly with
model size.

The most consistent gains occur on code-generation and knowledge-oriented tasks:
HumanEval, LiveCodeBench, and GPQA-Diamond improve for all six models, with the
largest gain observed on GPQA-Diamond for Qwen3-1.7B (+6.56 percentage points).
GSM8K and MMLU-Redux also improve for all six models. HumanEval+ improves for
five models and remains unchanged on Qwen3-8B, while TruthfulQA, PopQA, and BBH
do not decrease.

Overall, NGM provides gains across model families and scales, but the magnitude
of the improvement varies across tasks, with the clearest benefits appearing
on code- and knowledge-oriented benchmarks. These results support the
compatibility of the local-memory signal across the evaluated models without
updating backbone weights, adding trainable parameters, or using external
retrieval.

\subsection{Ablation Studies}
\label{sec:ablation}

\begin{table}[!htb]
\vspace{-5pt}
\caption{Component ablations on Qwen3-1.7B. The reference configuration uses
ReLU gating, stack fusion, and the standard tokenizer. Each variant changes one
component while keeping the remaining settings fixed.}
\label{tab:abl_components}
\centering
\small
\setlength{\tabcolsep}{3.5pt}
\begin{tabular}{l cccccc}
\toprule
Configuration & \textbf{GSM8K} & \textbf{MATH500} & \textbf{HE}
& \textbf{HE+} & \textbf{GPQA-D} & \textbf{TQA} \\
\midrule
Full NGM & \textbf{83.19} & 73.60 & \textbf{63.41}
& \textbf{59.73} & \textbf{33.33} & \textbf{52.51} \\
$-$ ReLU gating & 83.62 & 71.60 & \textbf{63.41}
& 58.54 & 30.81 & 52.39 \\
Alternative fusion (concat) & 82.34 & 72.60 & 62.80
& 59.15 & 27.78 & 52.14 \\
Compressed tokenization & 82.64 & \textbf{74.20} & 60.98
& 56.71 & 31.31 & 52.26 \\
\bottomrule
\end{tabular}
\vspace{-5pt}
\end{table}
\FloatBarrier

We study the key design choices of the Cosine-Gated Memory Injector and the
Causal N-gram Encoder on Qwen3-1.7B. The reference configuration uses
$\mathcal{N}=\{2,3\}$, ReLU gating, stack fusion, and the standard tokenizer.
All variants use the same output scales and insertion layers
(see Appendix Table~\ref{tab:appendix_ngm_settings}), while the backbone weights
remain frozen. HE and HE+ denote the pass@1 scores of HumanEval and HumanEval+,
respectively; GPQA-D denotes GPQA-Diamond, and TQA denotes TruthfulQA.
Table~\ref{tab:abl_components} reports three component variants. The full
configuration achieves the best or tied-best result on four of the six metrics,
indicating a favorable balance across tasks.

\paragraph{ReLU gating.} For the Cosine-Gated Memory Injector, removing ReLU improves GSM8K by
0.43 percentage points, but decreases MATH500 and GPQA-D by 2.00 and
2.52 points, respectively. This suggests that ReLU filters candidates whose
directions oppose the current decoder state, allowing the injector to focus on
positively aligned memory signals, although its effect is not uniformly
beneficial across tasks.

\paragraph{Alternative fusion.} For the Causal N-gram Encoder, stack fusion outperforms concat fusion on all six
metrics, with the largest difference on GPQA-D. Stack fusion computes separate
gates for different $N$-gram scales before aggregating the candidates, whereas
concat fusion uses a shared gate and mean aggregation.

\paragraph{Compressed tokenization. }We also introduce the compressed tokenization scheme from
Engram~\citep{cheng2026conditional} into the Causal N-gram Encoder. The
compressed token IDs are used to index the original embedding matrix, while the
backbone tokenizer and sequence length remain unchanged. Compressed tokenization
improves MATH500 by 0.60 points but performs worse than the standard tokenizer
on the other five metrics, resulting in a 1.28-point lower average across the
six tasks. This indicates that the Causal N-gram Encoder depends not only on the
surface form of tokens, but also on the geometric structure preserved by the
original token embeddings. Merging token IDs changes the local candidate
representations and consequently affects their alignment with the decoder state.

\begin{wraptable}{r}{0.72\columnwidth}
\vspace{-5pt}
\caption{Effect of $N$-gram window sizes on Qwen3-1.7B. All other components follow the reference configuration.}
\label{tab:abl_qwen17}
\centering
\small
\setlength{\tabcolsep}{4pt}
\begin{tabular}{l cccccc}
\toprule
$\mathcal{N}$ & \textbf{GSM8K} & \textbf{MATH500} & \textbf{HE} & \textbf{HE+} & \textbf{GPQA-D} & \textbf{TQA} \\
\midrule
$\{2\}$ & 82.64 & 72.60 & 61.59 & \textbf{60.37} & 30.30 & \textbf{52.63} \\
$\{3\}$ & 82.64 & 72.80 & 62.80 & 59.15 & 31.82 & 52.39 \\
$\{4\}$ & 83.09 & 73.40 & \textbf{63.41} & 58.54 & 30.81 & 52.14 \\
$\{2,3,4\}$ & 82.87 & 72.40 & 62.80 & 59.15 & 31.31 & 52.51 \\
\midrule
$\{2,3\}$ & \textbf{83.19} & \textbf{73.60} & \textbf{63.41} & 59.73 & \textbf{33.33} & 52.51 \\
\bottomrule
\end{tabular}
\vspace{-5pt}
\end{wraptable}
\FloatBarrier

\textbf{$N$-gram window sizes.}
Finally, we examine the window sizes used by the Causal N-gram Encoder.
Table~\ref{tab:abl_qwen17} shows that $\mathcal{N}=\{2,3\}$ achieves the highest
scores on GSM8K, MATH500, and GPQA-D, and ties for the best HumanEval score.
Single-scale configurations perform better on some individual metrics, such as
HumanEval+ or TruthfulQA, but adding a fourth scale does not provide a
consistent benefit. We therefore use $\mathcal{N}=\{2,3\}$ as the default
multi-scale configuration for capturing complementary local patterns.

\subsection{Extension to multimodal models}
\label{sec:multimodal}

\begin{table}[ht]
\caption{Preliminary multimodal results on Qwen3-VL-2B-Instruct. MMBench~\citep{liu2024mmbench}, MMStar~\citep{chen2024we}, OCRBench~\citep{liu2024ocrbench} (scored out of 1000), TQA = TruthfulQA~\citep{lin2022truthfulqa}, and MMLU-Redux~\citep{gema2025we}. \method{} is applied only to the language decoder; the visual encoder is unchanged.}
\label{tab:multimodal}
\centering
\small
\begin{tabular}{l ccccc}
\toprule
\textbf{Model} & \textbf{MMBench} & \textbf{MMStar} & \textbf{OCRBench} & \textbf{TQA} & \textbf{MMLU-R} \\
\midrule
Qwen3-VL-2B          & 76.32 & 54.67 & 826 & 52.26 & 67.79 \\
~~+\method{}          & \textbf{76.63} & \textbf{56.20} & \textbf{829} & \textbf{52.63} & \textbf{67.82} \\
\bottomrule
\end{tabular}
\end{table}

To test whether \method{} transfers beyond text-only LLMs, we apply \method{} to Qwen3-VL-2B-Instruct~\citep{bai2025qwen3}, leaving the visual encoder and vision-language fusion modules unchanged. The \ngram{} operates exclusively on text token embeddings; vision tokens are excluded from the sliding-window pooling so that the local memory signal remains purely linguistic.
Using VLMEvalKit~\citep{duan2024vlmevalkit} under identical decoding settings, \method{} improves or matches the baseline on all five reported benchmarks, with the largest gain on MMStar (+1.53; Table~\ref{tab:multimodal}). The retained TruthfulQA point estimates use a legacy evaluator whose metric has not been verified as MC2; they are not directly comparable with the generative TruthfulQA results in the main and ablation tables.
This single-scale result suggests that the same training-free local-memory mechanism can transfer to multimodal models without architectural changes, but we leave comprehensive multimodal evaluation to future work.

\subsection{Mechanistic analysis}
\label{sec:mechanistic}

We examine two mechanistic questions: whether the cosine gate reflects meaningful aligned structure rather than a generic embedding prior, and whether the resulting interactions are local in position.

\begin{figure*}[ht]
\centering
\includegraphics[width=0.65\textwidth]{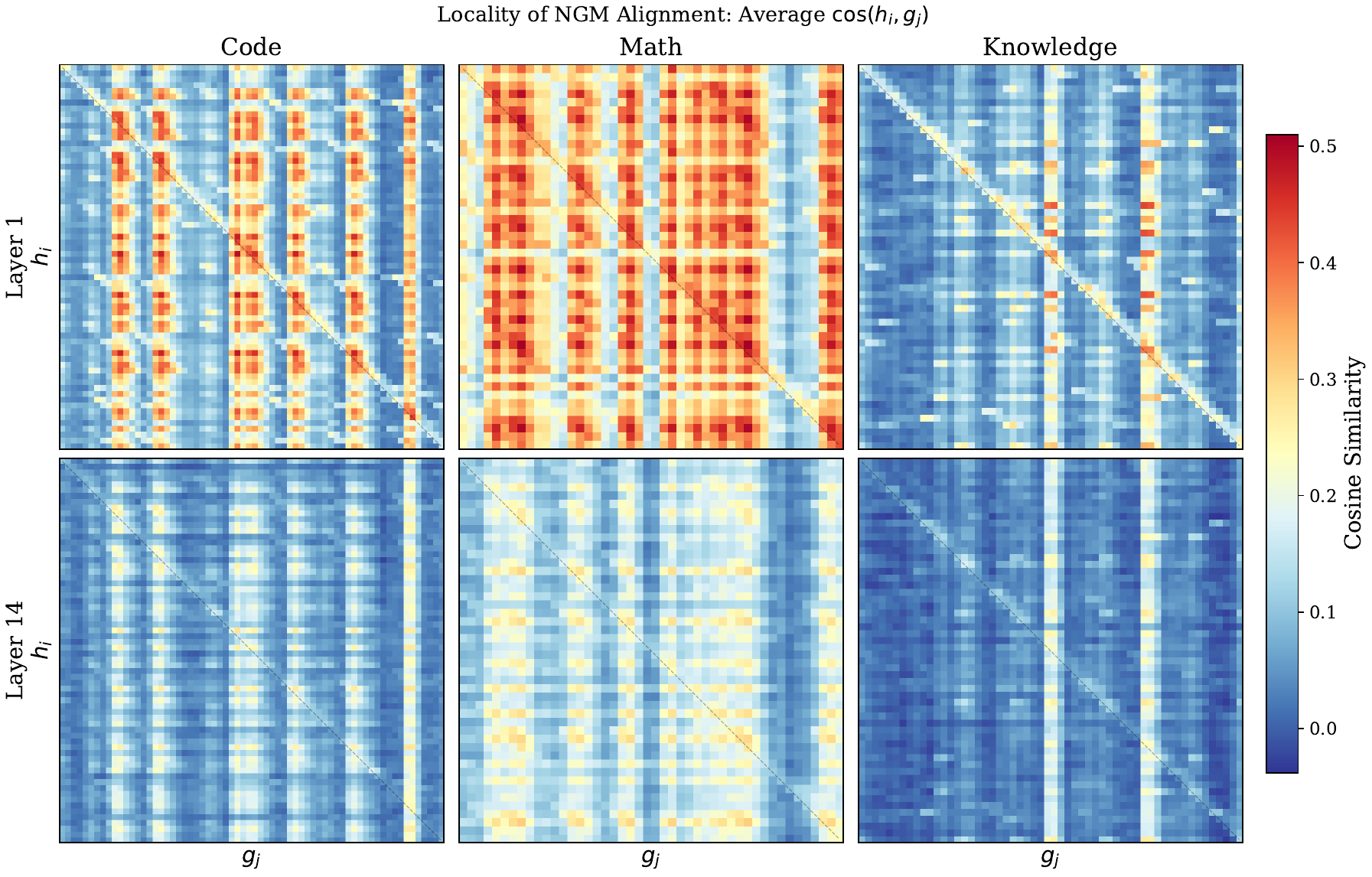}
\caption{Cross-position locality of \method{} interactions in the default Qwen3-8B-\method{} model. Heatmaps show the average cross-position cosine matrix $\cos(\boldsymbol{h}_i, \boldsymbol{g}_j)$ at the two default injection layers for representative code, math, and knowledge samples. The diagonal structure dominates, indicating that useful memory interactions are predominantly local.}
\label{fig:mech_locality}
\end{figure*}

\paragraph{Interactions are predominantly local.}
We next examine the full cross-position matrix $\cos(\boldsymbol{h}_i, \boldsymbol{g}_j)$ on the same model (Figure~\ref{fig:mech_locality}). Across representative code, math, and knowledge samples, the diagonal mean consistently exceeds the off-diagonal mean, with diagonal/off-diagonal ratios of 1.27$\times$--2.42$\times$ at the injected layers. The pattern is strongest for the knowledge sample and remains clear for code, indicating that the most useful memory signal is concentrated near the aligned position rather than uniformly distributed across the sequence.

\paragraph{Implication for the default gate.}
Together, these results support the default design. Since alignment remains well above shuffled and random controls, raw cosine similarity is informative without a learned projection. Since the interaction pattern is diagonal-dominant, the token-wise gate captures most of the useful signal while preserving linear-time cost. This interpretation is consistent with the strongest gains in Table~\ref{tab:main_results}, especially on tasks where short-range pattern stability matters. Combined with our intentionally restrictive evaluation setting---fixed backbone weights, no additional trainable parameters, and no external knowledge---this also yields a relatively controlled comparison in which the observed gains can be attributed more directly to the core \method{} mechanism. 

\label{app:wall_clock_overhead}
\subsection{Wall-clock overhead}
\label{sec:wall_clock_overhead}

\begin{wrapfigure}{r}{0.61\textwidth}
\centering
\includegraphics[width=\linewidth]{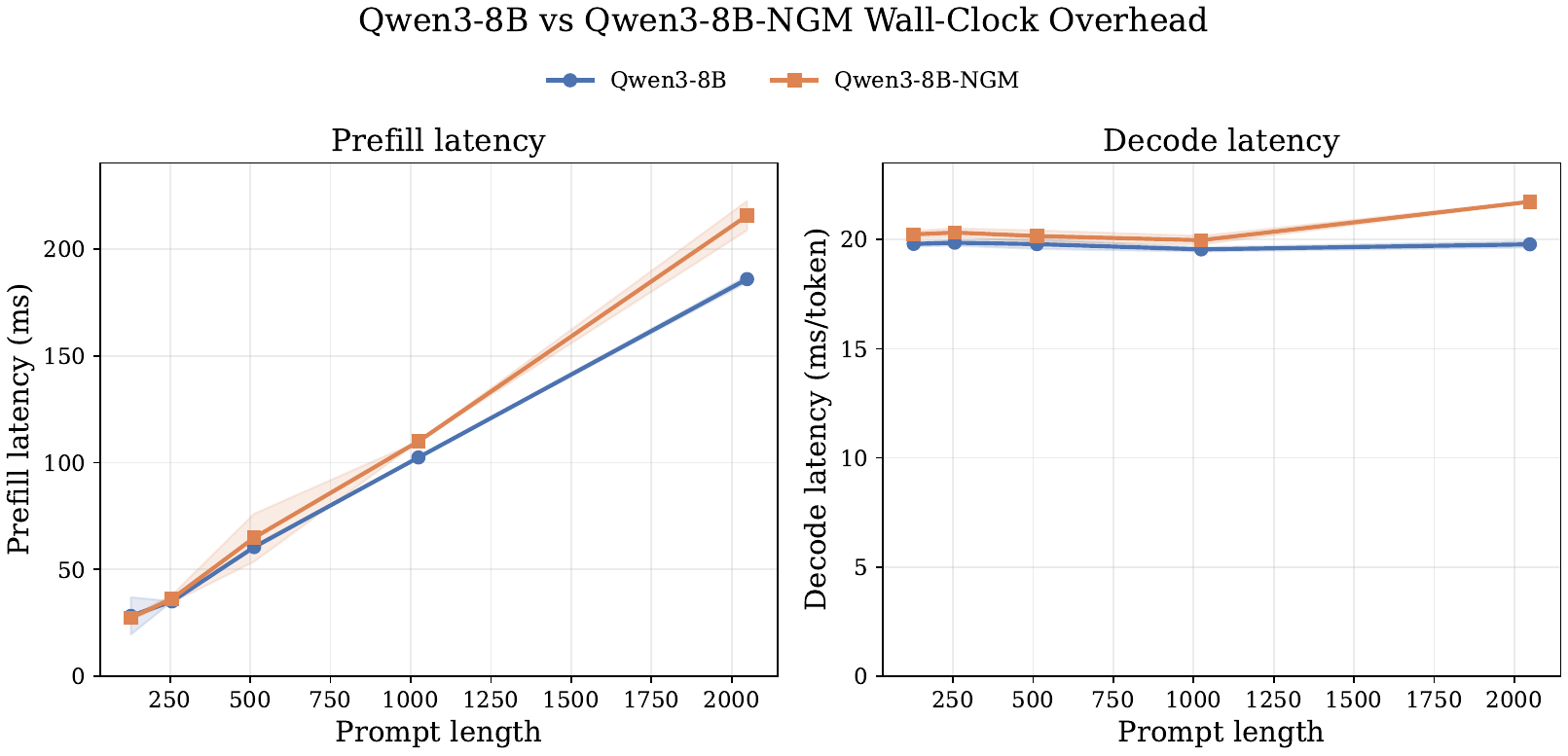}
\caption{Prefill and per-token decode latency for Qwen3-8B vs.\ Qwen3-8B-\method{} on a single RTX 5090 (mean $\pm$ std over 20 runs). The gap widens at 2048 tokens because the current implementation recomputes \ngram{} features over the full prefix.}
\label{fig:wall_clock_overhead}
\end{wrapfigure}

We measure prefill and decode latency for Qwen3-8B with and without \method{} on a single RTX 5090 (batch size 1, \texttt{bfloat16}, prompt lengths 128--2048, 20 runs; Figure~\ref{fig:wall_clock_overhead}).
Overhead is reported as the relative latency increase over the original Qwen3-8B baseline, i.e., $(\text{latency}_{\method{}}-\text{latency}_{\text{base}})/\text{latency}_{\text{base}}$.

For 256--1024-token prompts, prefill overhead is 3.4--7.3\% and decode overhead 1.9--2.3\%; at 2048 tokens the figures rise to 16.0\% and 9.9\%, respectively. These numbers reflect the released code path, which recomputes \ngram{} features over the full prefix; a streaming cache would reduce the cost to $O(|\mathcal{N}|d)$ per step.

\section{Conclusion}
\label{sec:conclusion}

We presented \method{}, a training-free and plug-and-play memory module built on the observation that pretrained token embeddings can serve as a non-parametric substrate for local memory. Rather than learning a separate memory representation, \method{} directly constructs causal multi-scale \ngram{} memories from the backbone embedding space and injects them through a non-parametric cosine gate, without modifying backbone parameters. Across multiple Qwen3 and Llama backbones, ten benchmarks, and a vision-language model, \method{} consistently improves performance across model families and scales, with notable gains in code generation and complex reasoning. These results suggest that useful local lexical and compositional structure is already encoded in pretrained representation spaces and can be directly reused as memory. More broadly, our findings motivate treating pretrained representations not only as inputs to computation, but also as reusable computational resources for augmenting language models without additional training.

\clearpage

\section*{AI Use Statement}

Generative AI tools were used solely to assist with language
editing and wording refinement during manuscript preparation.
The authors reviewed the AI-assisted text and take full
responsibility for the final content of this work.

\section*{Ethics Statement}

This work studies inference-time augmentation of pretrained
language models using existing model checkpoints and evaluation
benchmarks. The resulting models may retain biases, factual
errors, or other undesirable behaviors of the underlying models.
Our evaluation focuses on task performance and does not establish
improvements in safety or fairness. The reported benchmark gains
should therefore not be interpreted as guarantees of reliability
or suitability for sensitive applications. Downstream use should
consider these limitations and respect the licenses and usage
restrictions of the underlying models and datasets.

\section*{Reproducibility Statement}

We describe the Causal N-gram Encoder and Cosine-Gated Memory
Injector in Section~\ref{sec:method}, with the overall procedure
summarized in Algorithm~\ref{app:ngm_code}.
The evaluated backbones, benchmarks, and inference settings
are described in Section~\ref{sec:experiments}.
Appendix~\ref{app:ngm_code} provides a simplified PyTorch
implementation, and Table~\ref{tab:appendix_ngm_settings}
reports the output scales and insertion layers for each backbone.
The hardware configuration and measurement protocol for the
latency evaluation are specified in
Section~\ref{sec:wall_clock_overhead}.

\bibliographystyle{iclr2027_conference}
\bibliography{custom}

\clearpage
\appendix
\section{\method{} implementation}
\label{app:ngm_code}

Listing~\ref{lst:ngm_code} gives a simplified PyTorch implementation of \method{}.

\begin{lstlisting}[style=pypseudo, caption={\method{} core implementation (simplified).}, label={lst:ngm_code}]
def ngm_forward(hidden_states, input_ids, embed_matrix,
                ngram_sizes, output_scale, use_relu):
    """
    hidden_states: [B, L, D]   (layer-l hidden states)
    input_ids:     [B, T]      (full input token ids)
    embed_matrix:  Embedding    (token embeddings from backbone model)
    """
    B, L, D = hidden_states.shape
    T = input_ids.shape[1]

    # Step 1: causal n-gram construction
    token_emb = embed_matrix(input_ids)          # [B, T, D]
    ngram_list = []
    for n in ngram_sizes:
        padded = F.pad(token_emb, (0, 0, n-1, 0))
        pooled = F.avg_pool1d(
            padded.transpose(1, 2), kernel_size=n, stride=1
        ).transpose(1, 2)                        # [B, T, D]
        ngram_list.append(pooled)
    ngram_emb = torch.stack(ngram_list, dim=2)   # [B, T, N, D]

    # Step 2: KV-cache alignment
    ngram_emb = ngram_emb[:, -L:, :, :]          # [B, L, N, D]

    # Step 3: cosine gating
    h_norm = F.normalize(hidden_states, dim=-1)  # [B, L, D]
    g_norm = F.normalize(ngram_emb, dim=-1)      # [B, L, N, D]
    sim = torch.einsum('bld,blnd->bln', h_norm, g_norm)
    if use_relu:
        sim = F.relu(sim)

    # Step 4: residual injection
    out = torch.einsum('bln,blnd->bld', sim, ngram_emb)
    return hidden_states + output_scale * out
\end{lstlisting}

\subsection{Model-specific \method{} settings}

The main evaluation uses $\mathcal{N}=\{2,3\}$, ReLU gating, and stack fusion. The output scale is fixed at $0.1$. Every scale remains frozen during evaluation. Table~\ref{tab:appendix_ngm_settings} reports inserted decoder layers in 1-based numbering; the implementation uses zero-based indices $\{1,14\}$.

\begin{table}[ht]
\caption{NGM settings for the six backbones in the main evaluation.}
\label{tab:appendix_ngm_settings}
\centering
\small
\begin{tabular}{lccc}
\toprule
\textbf{Model} & \textbf{Backbone layers} & \textbf{Inserted decoder layers} & \textbf{Output scale} \\
\midrule
Qwen3-0.6B-\method{} & 28 & 2, 15 & 0.1 \\
Qwen3-1.7B-\method{} & 28 & 2, 15 & 0.1 \\
Qwen3-4B-\method{} & 36 & 2, 15 & 0.1 \\
Qwen3-8B-\method{} & 36 & 2, 15 & 0.1 \\
\midrule
Llama-1B-\method{} & 16 & 2, 15 & 0.1 \\
Llama-3B-\method{} & 28 & 2, 15 & 0.1 \\
\bottomrule
\end{tabular}
\end{table}

These layer placements are heuristic. Engram's layer-sensitivity analysis~\citep{cheng2026conditional} motivates balancing early access to local features with enough contextualization for gating. We use the 2nd and 15th decoder layers as a shared practical default across the six backbones; their relative depths differ, especially for 16-layer Llama-1B, so the shared indices are not equivalent relative positions or universally optimal placements.

\section{Case studies: \method{} vs.\ base model on GSM8K}
\label{app:case_studies}

We present three representative examples from the GSM8K benchmark where Qwen3-8B-\method{} produces the correct answer while the base Qwen3-8B model fails.
Each case highlights a different failure mode that \method{} helps mitigate.

\begin{tcolorbox}[colback=gray!4, colframe=black!60, boxrule=0.5pt,
  title={\textbf{Case 1: Arithmetic error correction}}, fonttitle=\small,
  coltitle=black, colbacktitle=gray!15]
\textbf{Question.}\;
In a candy machine, there are 22 more than four times the number of pink gumballs as there are blue gumballs. If there are 12 blue gumballs, how many pink ones are there?
\hfill \textbf{Gold answer:} 70
\tcblower
\noindent\begin{minipage}[t]{0.48\textwidth}
\textbf{Base model (Qwen3-8B) \ding{55}}\\[4pt]
The number of pink gumballs is 22 more than four times the number of blue gumballs.\\
Given that there are 12 blue gumballs, we calculate:\\
$4 \times 12 = \boldsymbol{28}$\quad {\color{red}(arithmetic error)}\\
$28 + 22 = 50$\\[2pt]
\texttt{ANSWER: 50}
\end{minipage}\hfill{\color{black!30}\vrule width 0.4pt}\hfill\begin{minipage}[t]{0.47\textwidth}
\textbf{Qwen3-8B-\method{} \ding{51}}\\[4pt]
The number of pink gumballs is 22 more than four times the number of blue gumballs.\\
There are 12 blue gumballs.\\
$4 \times 12 = \boldsymbol{48}$\quad {\color{teal}(correct)}\\
$48 + 22 = 70$\\[2pt]
\texttt{ANSWER: 70}
\end{minipage}

\tcbline
\noindent\textbf{\textit{Analysis.}}
The base model produces an incorrect intermediate product ($4\times 12=28$), causing the final answer to be wrong.
\method{} maintains correct arithmetic in the short-range computation chain.
\end{tcolorbox}

\begin{tcolorbox}[colback=gray!4, colframe=black!60, boxrule=0.5pt,
  title={\textbf{Case 2: Semantic misinterpretation}}, fonttitle=\small,
  coltitle=black, colbacktitle=gray!15]
\textbf{Question.}\;
Finn watches 11 tadpoles swimming in the pond. Suddenly he sees 6 of them come out of hiding from under a lily pad, then he sees 2 of them hide under a rock. How many tadpoles can Finn see in the pond now?
\hfill \textbf{Gold answer:} 15
\tcblower
\noindent\begin{minipage}[t]{0.48\textwidth}
\textbf{Base model (Qwen3-8B) \ding{55}}\\[4pt]
Finn initially sees 11 tadpoles.\\
Then, 6 come out of hiding from under a lily pad.\\
After that, 2 hide under a rock.\\[2pt]
{\color{red}Ignores the 6 newcomers:}\\
$11 - 2 = 9$\\[2pt]
\texttt{ANSWER: 9}
\end{minipage}\hfill{\color{black!30}\vrule width 0.4pt}\hfill\begin{minipage}[t]{0.47\textwidth}
\textbf{Qwen3-8B-\method{} \ding{51}}\\[4pt]
Finn initially watches 11 tadpoles.\\
Then, 6 come out of hiding $\rightarrow$ visible count increases by 6.\\
Next, 2 hide under a rock $\rightarrow$ visible count decreases by 2.\\[2pt]
$11 + 6 - 2 = 15$\\[2pt]
\texttt{ANSWER: 15}
\end{minipage}

\tcbline
\noindent\textbf{\textit{Analysis.}}
The base model misinterprets ``come out of hiding'' and ignores the additional 6 tadpoles, computing only $11-2$.
\method{} correctly accumulates all quantity changes mentioned in the prompt.
\end{tcolorbox}

\newpage
\begin{tcolorbox}[enhanced, breakable, colback=gray!4, colframe=black!60, boxrule=0.5pt,
  title={\textbf{Case 3: Reading comprehension error}}, fonttitle=\small,
  coltitle=black, colbacktitle=gray!15]
\textbf{Question.}\;
Three friends spent \$20.25 on 3 tickets to the fair. They also spent \$4.50 less on food than on the tickets. They also went on 2 different rides which cost \$33 for each ride. If they agreed to split all the costs evenly, how much did each of them pay?
\hfill \textbf{Gold answer:} 34
\tcblower
\noindent\begin{minipage}[t]{0.48\textwidth}
\textbf{Base model (Qwen3-8B) \ding{55}}\\[4pt]
{\color{red}Misreads \$20.25 as per-ticket price:}\\
Tickets: $3\times\$20.25=\$60.75$\\
Food: $\$60.75 - \$4.50 = \$56.25$\\
Rides: $2\times\$33 = \$66$\\
Total: $\$60.75+\$56.25+\$66=\$183$\\
Each: $\$183 \div 3 = \$61$\\[2pt]
\texttt{ANSWER: 61}
\end{minipage}\hfill{\color{black!30}\vrule width 0.4pt}\hfill\begin{minipage}[t]{0.47\textwidth}
\textbf{Qwen3-8B-\method{} \ding{51}}\\[4pt]
Tickets: \$20.25 (total for 3)\\
Food: $\$20.25 - \$4.50 = \$15.75$\\
Rides: $2\times\$33 = \$66$\\
Total: $\$20.25+\$15.75+\$66=\$102$\\
Each: $\$102 \div 3 = \$34$\\[2pt]
\texttt{ANSWER: 34}
\end{minipage}

\tcbline
\noindent\textbf{\textit{Analysis.}}
The base model misreads ``spent \$20.25 on 3 tickets'' as \$20.25 \emph{per ticket}, inflating all downstream totals.
\method{} correctly interprets \$20.25 as the total ticket expenditure, preserving local textual coherence across the reasoning chain.
\end{tcolorbox}

\end{document}